# Automated Batch-Variable Selection for ComBat-Based Radiomic Harmonization

Jose Lozano-Montoya[1,2,*], Gemma Urbanos[2], Carmen Prieto-de-la-Lastra[2], Emilio Soria-Olivas[3], Ana Jimenez-Pastor[2]

[1] Universitat de València, Valencia, Spain

[2] Research & Frontier in AI Department, Quantitative Imaging Biomarkers in Medicine, Valencia, Spain

[3] Intelligent Data Analysis Laboratory, Universitat de València, Valencia, Spain

* Corresponding author: Jose Lozano-Montoya (joselozano@quibim.com)

## Abstract

### Objectives

ComBat harmonizes multicentre radiomic features, but batch variables are usually chosen manually, without a standardized criterion for selecting among image acquisition parameters. We propose ParsiComBat, which selects the most suitable acquisition parameters for ComBat, balancing batch-effect removal against feature-structure preservation, abstaining when technical variation is negligible.

### Materials and methods

ParsiComBat ranks candidate image acquisition parameters by batch effect, filters redundancy, and scores sequential ComBat corrections. We evaluated it in three experiments: in simulations, testing whether it recovered a known batch variable and abstained without one; in CT (NSCLC-Radiomics) and MRI (MAMA-MIA) cohorts, comparing its batch-effect removal against correction for manufacturer (manual), for every feasible parameter (greedy) and AutoCombat; and in clinical prediction, assessing each strategy internally and externally.

### Results

In simulations, ParsiComBat identified the true batch variable in 92.9% of positive trials and abstained in 99.2% of null controls. In CT, correcting for reconstruction kernel reduced mean k-nearest neighbour batch-effect test from 0.276 to 0.032, almost matching greedy correction of six variables. In MRI, correcting for slice thickness and manufacturer reduced it from 0.680 to 0.325, slightly more than manual selection but less than greedy correction of seven variables. Removing technical variation did not improve clinical discrimination: no strategy, including raw features, differed significantly from ParsiComBat in area under the curve.

### Conclusion

ParsiComBat provides a data-driven framework that removes technical batch effects using minimal acquisition parameters, preserving radiomic feature structure and clinical model discrimination. It selects variables verifiable against the acquisition protocol and abstains when no correction is needed, supporting transparent and reproducible multicentre radiomics.

## Key points

- **Question:** Can automated selection of image acquisition parameters optimize ComBat radiomic harmonization and improve downstream clinical model discrimination across scanners and centres?
- **Findings:** Automated selection recovered simulated batch variables and reduced technical variation with one or two acquisition parameters, without compromising clinical model discrimination.
- **Relevance statement:** Radiomic studies should report which batch variables they corrected for, and harmonize only when the data show a batch effect.

## Abbreviations

| Abbreviation | Meaning |
|---|---|
| ASW | Average silhouette width |
| AUC | Area under the curve |
| CI | Confidence interval |
| CV | Cross-validation |
| DCE | Dynamic contrast-enhanced |
| kBET | k-nearest neighbour batch-effect test |
| NSCLC | Non-small cell lung cancer |
| pCR | Pathological complete response |
| PERMANOVA | Permutational multivariate analysis of variance |
| RESI | Robust effect size index |
| ROC | Receiver operating characteristic |

## Introduction

Radiomics extracts quantitative features from medical images to build diagnostic, prognostic and treatment-response models [1,2]. However, feature values also depend on acquisition settings such as scanner model, manufacturer, CT reconstruction kernel or MRI field strength [3–5]. When images from several scanners or centres are pooled, this technical variation, or batch effect, can mask or mimic biological differences and limit generalizability [6,7].

Harmonization can operate directly on the images [8,9] or on the extracted features [10], where ComBat, an empirical Bayes method from genomics, is widely used [11–14]. It removes differences in the location and scale of each feature between the levels of a batch variable, such as centre or manufacturer. Technical variation, however, arises from several related acquisition parameters whose effects differ [4,5], and choosing among them can change the performance of the resulting models [15]. Correcting for one obvious variable may miss the dominant batch effect, whereas correcting for every recorded parameter may cause overcorrection. In addition, applying ComBat on a variable without a batch effect can obscure true biological effects [16] and introduce spurious ones, increasingly so as more variables are corrected [17].

Among ComBat extensions, Nested ComBat corrects several batch variables sequentially, and OPNested ComBat optimizes their order [18], but both adjust for every variable provided. Gaussian mixture variants and AutoCombat instead infer batch labels, by splitting feature distributions or clustering acquisition parameters [19,20]. None selects among recorded image variables or decides whether harmonization is needed.

We propose ParsiComBat, which selects one or more image acquisition parameters as ComBat batch variables based on the batch effect removed and the feature structure preserved, and abstains when no effect is large enough. We evaluated it in simulated data and in CT and MRI cohorts, measuring batch-effect removal and, because the effect of harmonization on downstream models has been inconsistent [18,21–23], the discrimination of clinical prediction models.

## Materials and Methods

### Study design

This retrospective study used publicly available, de-identified data from studies that obtained ethics approval and, where required, informed consent. Institutional review board approval and informed consent were therefore not required for this secondary analysis. ParsiComBat was evaluated across three experiments (Figure 1). Reporting follows the CLEAR checklist [24] (see Supplementary Material).

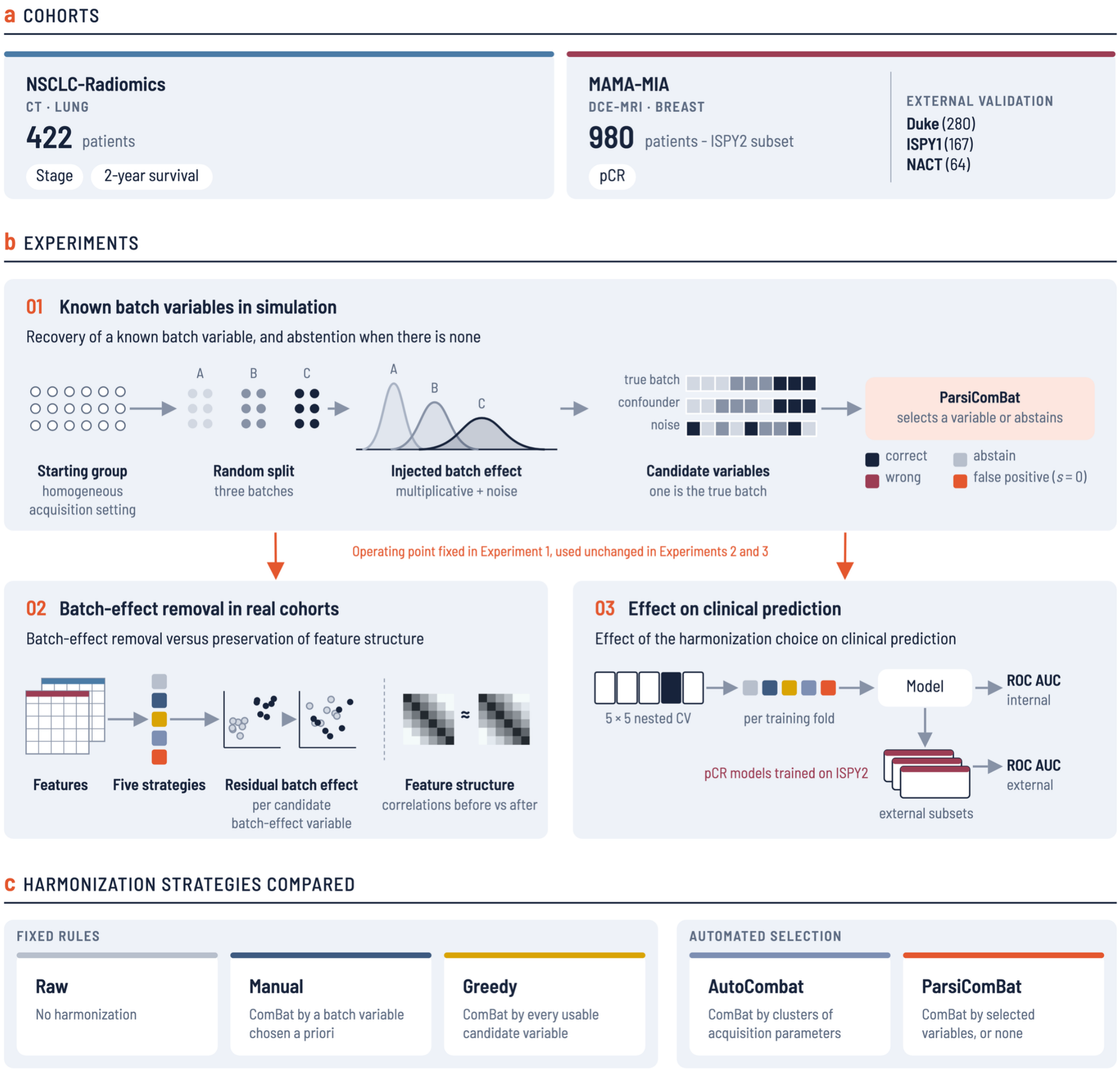

**Figure 1.** Study design. (a) Cohorts used in the study, with different imaging modalities, number of patients and endpoints. (b) The three experiments: Experiment 1 injects a known batch effect into groups scanned under one acquisition setting and fixes the operating point of ParsiComBat; Experiment 2 measures the residual batch effect and the preservation of feature structure in the real cohorts; Experiment 3 measures internal and external clinical prediction performance. (c) Harmonization strategies compared: fixed rules apply the same correction regardless of the data, while automated methods decide what to correct from the data; their colours identify them in panel **b**. pCR, pathological complete response; $n$, search depth of ParsiComBat; $s$, effect scale; CV, cross-validation.

## Data

### *Cohorts*

Two public collections were analysed (Table 1), including all patients with tumour segmentation and radiomic features. NSCLC-Radiomics, from The Cancer Imaging Archive, contains CT images of patients with non-small cell lung cancer (NSCLC) treated at a single centre and acquired with several scanners [1]. Its two endpoints were overall stage IIIa–IIIb versus I–II and death within two years of treatment start. MAMA-MIA is a multicentre collection of dynamic contrast-enhanced (DCE) breast MRI [25]. Its ISPY2 subset was used in all three experiments, and the Duke, ISPY1 and NACT subsets only for external validation, with pathological complete response (pCR) after neoadjuvant therapy as the endpoint. A patient was excluded from an endpoint when its label was missing or, for 2-year survival, when follow-up was censored before two years.

**Table 1.** Datasets analysed with patient characteristics, imaging modality, clinical endpoints and radiomic features.

| Collection | Dataset | Modality | Patients, n | Age (years), median (IQR) | Sex (male/female), n | Endpoint (positive/negative) | Radiomic features, n |
|---|---|---|---|---|---|---|---|
| NSCLC-Radiomics | NSCLC-Radiomics | CT | 422 | 68.6 (61.2–75.8) | 290/132 | Stage III vs I–II (288/133)<br>2-year survival (251 deaths/169 alive) | 1,413 |
| MAMA-MIA | ISPY2 | DCE-MRI | 980 | 49.0 (41.0–56.0) | 0/980 | pCR (316/664) | 1,365 |
| MAMA-MIA | Duke | DCE-MRI | 280 | 47.0 (41.8–55.0) | 0/280 | pCR (64/216) | 1,365 |
| MAMA-MIA | ISPY1 | DCE-MRI | 167 | 48.0 (41.5–53.5) | 0/167 | pCR (49/118) | 1,365 |
| MAMA-MIA | NACT | DCE-MRI | 64 | 48.0 (40.0–54.2) | 0/64 | pCR (11/53) | 1,365 |

CT, computed tomography; DCE-MRI, dynamic contrast-enhanced magnetic resonance imaging; IQR, interquartile range; NSCLC, non-small cell lung cancer; pCR, pathological complete response.

### *Radiomic features and candidate batch-effect variables*

For NSCLC-Radiomics, PyRadiomics features computed in the gross tumour volume were obtained from the Open-radiomics release [26,27]; for MAMA-MIA, they were extracted with PyRadiomics 3.0.1 from the first post-contrast phase within the expert segmentation. Shape features were never harmonized and constant features were removed before any subset was formed. The candidate batch-effect variables were the acquisition parameters recorded in the image metadata that were not constant, seven in CT (manufacturer, scanner model, peak tube voltage, tube current, exposure, exposure time and reconstruction kernel) and eight in MRI (field strength, pixel spacing, slice thickness, manufacturer, scanner model, bit depth, echo time and repetition time). Their distribution and processing are described in Supplementary Methods.

## ParsiComBat: automated selection of batch variables

ParsiComBat takes a radiomic feature matrix, a list of candidate batch-effect variables and, optionally, biological covariates whose effects ComBat must preserve. It returns an ordered list of batch variables for sequential ComBat, or it abstains and leaves the features unchanged (Figure 2).

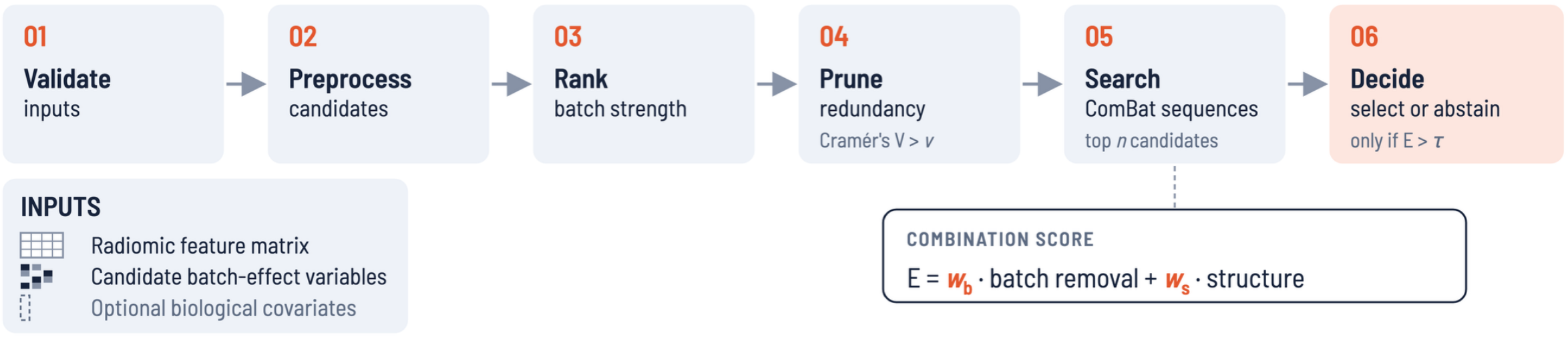

**Figure 2.** ParsiComBat workflow. Candidates are validated and preprocessed, ranked by batch strength, filtered for redundancy and combined into ComBat sequences that are scored after harmonization. The best sequence is used only if its score exceeds the threshold; otherwise, the selector abstains. Parameters: $v$, redundancy threshold; $n$, search depth; $w_b$ and $w_s$, weights of the batch and structure terms; τ, decision threshold.

After input checks (phase 1), candidates with more than ten distinct values are discretized at tertiles, and those with missing values, one group or any group under five patients are discarded (phase 2). The remaining candidates are ranked on the unharmonized features by a fixed weighted sum (0.4, 0.4 and 0.2) of three normalized measures of batch effect: the permutational multivariate analysis of variance (PERMANOVA) pseudo-F statistic [28], the robust effect size index (RESI) pooled over features [29] and the average silhouette width (ASW) [30]; the first two have been proposed to measure batch effects in radiomic features [31] (phase 3). Of each pair of candidates whose Cramér's V exceeds the redundancy threshold $v$, the lower-ranked one is removed (phase 4). The $n$ highest-ranked candidates are then combined into every ordered sequence of one to $n$ variables, and each sequence is applied with sequential ComBat [11,12,19] and scored (phase 5) with

$$E = w_b\,B + w_s\,S, \qquad (1)$$

where the batch term $B$ combines the relative reductions of the three measures of batch-effect removal and the structure term $S = (\rho + 1)/2$, with ρ being the Spearman correlation between the feature–feature correlation matrices before and after harmonization. If the best score exceeds the threshold τ, the variables of that sequence are returned; otherwise, the selector abstains (phase 6). Definitions and parameter values are in Supplementary Methods.

## Experiment 1: selection of known batch variables in simulated data

Experiment 1 injected a batch effect of known origin and strength into real features, to test whether the selector returns the true batch variable when an effect is present and abstains when there is none, and to fix the operating point for the remaining experiments. To start from data without a batch effect of their own, each simulation used a group of patients acquired with the most homogeneous acquisition settings available (Table 2).

**Table 2.** Acquisition groups used as starting feature matrices in Experiment 1. Patients within each NSCLC group share manufacturer, scanner model and reconstruction kernel. Patients within each MAMA-MIA group share manufacturer, scanner model (value given in parentheses), field strength and bit depth (DICOM high-bit value). Other parameters could vary within a group.

| Group | Patients, n | Manufacturer | Scanner model | Other fixed parameters |
|---|---|---|---|---|
| **NSCLC-Radiomics (CT)** | | | | |
| L1 | 142 | Siemens | Biograph 40 | kernel G1<br>(B10f, B18f, B19f) |

| Group | Patients, n | Manufacturer | Scanner model | Other fixed parameters |
|---|---|---|---|---|
| L2 | 55 | Siemens | Sensation Open | kernel G2 (B30f, B31f) |
| L3 | 98 | CMS | XiO | kernel not recorded |
| **MAMA-MIA, ISPY2 (DCE-MRI)** | | | | |
| B1 | 398 | GE | G4 (Signa Excite, Signa HDx, Signa HDxt) | field strength 1.5 T<br>bit depth 15 |
| B2 | 148 | GE | G4 (Signa Excite, Signa HDx, Signa HDxt) | field strength 3.0 T<br>bit depth 15 |
| B3 | 83 | Philips | G6 (Intera, Achieva) | field strength 1.5 T<br>bit depth 11 |
| B4 | 189 | Siemens | G1 (Avanto, SymphonyTim, Espree, Symphony) | field strength 1.5 T<br>bit depth 11 |

B1–B4, acquisition groups of the MAMA-MIA (ISPY2) cohort; L1–L3, acquisition groups of the NSCLC-Radiomics cohort; CT, computed tomography; DCE-MRI, dynamic contrast-enhanced magnetic resonance imaging; NSCLC, non-small cell lung cancer.

In each trial, the patients were assigned at random to three batches of near-equal size, and a batch effect with additive noise was injected into every non-shape feature:

$$x'_{ij} = (1 + c_{g(i)}\, s)\, x_{ij} + \eta\, \sigma_j\, \varepsilon_{ij}, \qquad c_g \in \{0,1,2\}, \qquad (2)$$

where $x_{ij}$ and $x'_{ij}$ are the original and perturbed values of feature $j$ for patient $i$, $s$ is the effect scale, $g(i)$ the batch of patient $i$, $\sigma_j$ the standard deviation of feature $j$, $\eta$ the noise level and $\varepsilon_{ij}$ standard normal noise.

The selector chose among the true batch labels, a confounder associated with them and a noise variable with random labels, see Figure 1b. The scenarios combined six effect scales ($s$ = 0, 0.005, 0.01, 0.02, 0.05 and 0.1), five noise levels ($\eta$ = 0, 0.1, 0.5, 1 and 2) and two confounder strengths (0.6 and 0.8), giving 60 scenarios per group. Each scenario was repeated 50 times with new random batches. Trials with scale 0 had no batch effect and served as null controls.

The primary outcomes were accuracy, the proportion of trials with a batch effect in which the true variable was returned, and the false-positive rate, the proportion of null controls with any selection. These outcomes were used to fix the batch weight $w_b$ and the threshold $\tau$. Forty configurations were compared: $w_b \in \{0.6, 0.7, 0.8, 0.9\}$ with $w_s = 1 - w_b$, and $\tau$ from 0.40 to 0.85 in steps of 0.05. Among the configurations whose false-positive rate did not exceed 5% at any noise level, the most accurate became the operating point.

### Experiment 2: batch-effect removal in real cohorts

Experiment 2 compared five strategies in all patients of both cohorts (Figure 1c):

- **Raw**: no harmonization (reference).
- **Manual**: ComBat on scanner manufacturer, a major source of variability in radiomic features [3,32] that gives few and large batches, as ComBat requires [14].
- **Greedy**: sequential ComBat on every candidate acquisition parameter that is feasible for ComBat, without selection.
- **AutoCombat**: ComBat on clusters of acquisition parameters [20], the closest published automated method.

- **ParsiComBat**: ComBat at the operating point of Experiment 1, with $n = 3$.

No biological covariates were used, except in a sensitivity analysis with sex protected in NSCLC. The residual batch effect of every candidate batch-effect variable was measured with RESI, pseudo-F, ASW and the k-nearest neighbour batch-effect test (kBET) rejection rate [33]; lower values mean a smaller residual effect. The selector uses the first three, so only kBET is an independent measure. Variables with residual PERMANOVA effects ($p < 0.05$) were counted, and feature-structure preservation was measured by pre–post harmonization Spearman correlation.

### Experiment 3: effect of harmonization on clinical prediction

Experiment 3 evaluated how harmonization affected the prediction of the three clinical endpoints using radiomic and clinical variables. Only non-shape radiomic features were harmonized. For each harmonization strategy, performance was estimated by nested cross-validation (CV) with five outer and five inner stratified folds. Discretization, ComBat and the AutoCombat clusters were fitted on each training fold and applied to its test fold. Inner folds selected the model with highest mean receiver operating characteristic (ROC) area under the curve (AUC) and set the classification threshold via Youden's J index. Five model families were evaluated: logistic regression, random forest, support vector classifier, XGBoost [34] and CatBoost [35]. The pCR models were trained only on ISPY2 and validated externally on the Duke, ISPY1 and NACT cohorts. External predictions averaged probabilities across the five outer-fold models. ROC AUC was the primary metric, and balanced accuracy and F1 score were secondary.

### Statistical analysis

Proportions are reported with Wilson 95% confidence intervals [36]. In Experiment 3, internal performance was the unweighted mean over the five outer folds and external performance was computed on all patients of each external cohort, with 95% confidence intervals from a paired, class-stratified bootstrap. ParsiComBat was compared with each other strategy using paired randomization tests, with Holm adjustment within each endpoint or cohort. Two analyses were added post hoc: accuracy in Experiment 1 was described against the RESI of the true batch variable before harmonization, and the radiomic features of each final internal model in Experiment 3 were counted.

### Use of AI tools

A large language model-based assistant (Claude, Anthropic) was used to support software development, code review, literature search and language editing. The authors reviewed and verified all outputs and take full responsibility for the content of this article.

## Results

### Experiment 1: selection of known batch variables in simulated data

ParsiComBat found the variable that caused the batch effect in most trials and almost never harmonized when there was no effect to remove. Of the 21,000 trials, 17,500 contained a simulated batch effect and 3,500 were null controls. The selector returned the true batch variable in 92.9% (95% CI 92.5–93.3%) of trials with a batch effect. In the null controls, it wrongly selected a variable in only 0.8% (95% CI 0.5–1.1%) of trials.

These settings were chosen as shown in Figure 3a. We required the selector to harmonize in no more than 5% of null controls at every noise level. Of the 40 configurations, 18 met this limit, and the most accurate became the operating point (Figure 3a; $w_b$ = 0.9, $w_s$ = 0.1, $\tau$ = 0.60).

Whether the selector detected a batch effect depended mainly on how large the effect was (Figure 3b). We measured this size with the RESI of the true batch variable, where 0 means no difference between batches and larger values mean larger differences. Below a RESI of 0.20, the selector almost always abstained; from 0.30 upwards, it returned the true variable in 99.7% of trials. This range matters for the real cohorts: before harmonization, their candidate batch-effect variables had a RESI of 0.04–0.25 in NSCLC and 0.10–0.35 in MAMA-MIA, near or below this transition.

When the selector missed the true batch variable, it almost always abstained rather than choosing a wrong one (Figure 3c): of 1,236 misses, 1,232 were abstentions and 4 were wrong selections. A miss therefore left the features unharmonized instead of harmonizing them by the wrong variable. Almost all misses occurred in MAMA-MIA at the two weakest effect scales. Outcomes by noise level and starting group are in Supplementary Results.

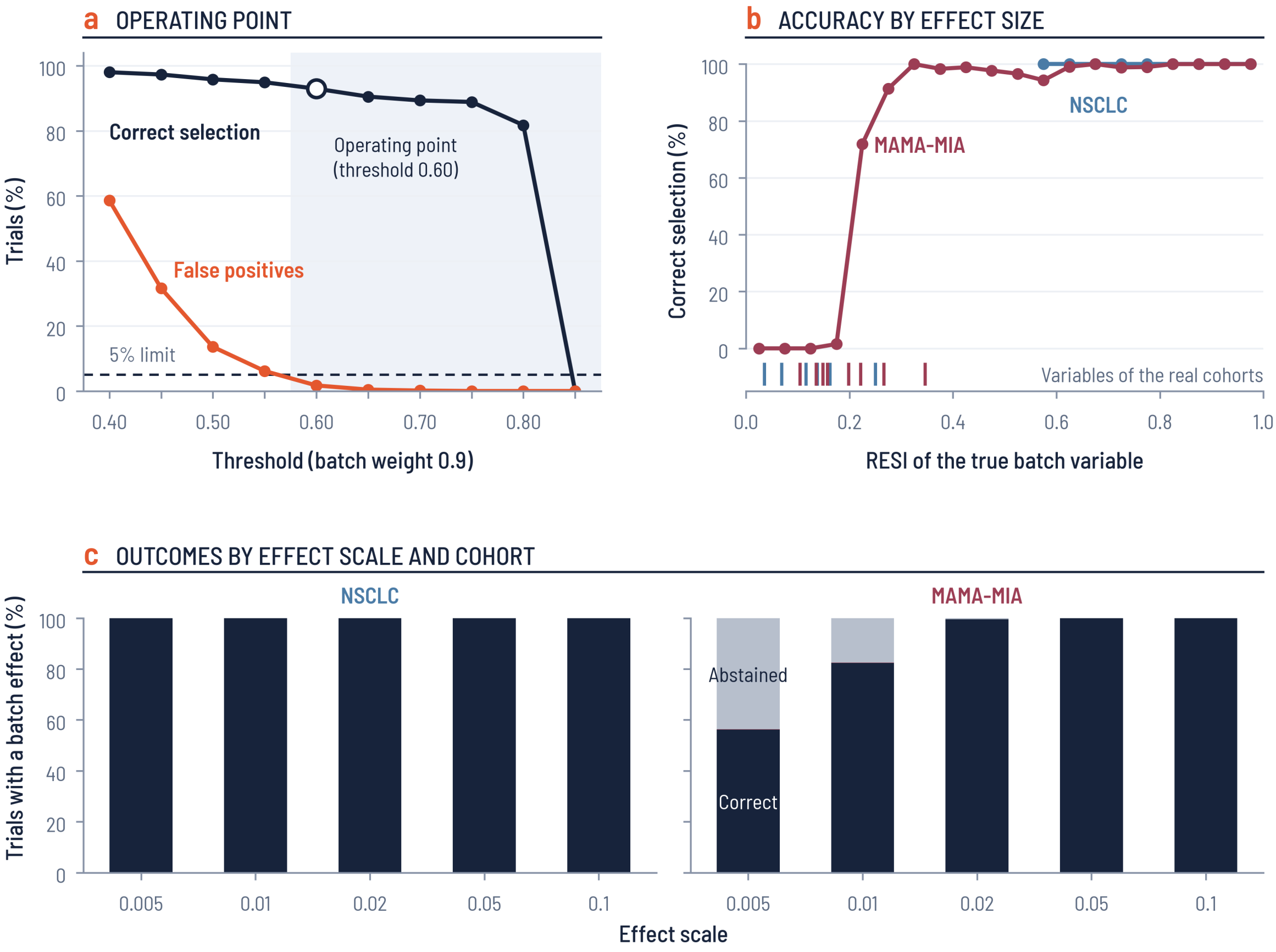

**Figure 3.** Results of Experiment 1. (a) Choice of the operating point, for $w_b$ = 0.9. For each threshold $\tau$, the black line gives the percentage of trials with a batch effect in which the true batch variable was selected, and the orange line the percentage of null controls in which any variable was selected, at the noise level where this percentage was highest. The dashed line is the 5% limit; thresholds in the shaded band meet it at every noise level, and the open circle marks the selected operating point. (b) Accuracy at the operating point against the size of the batch effect in each trial, measured as the RESI of the true batch variable before harmonization. Each point is a bin of 0.05 with at least 20 trials, coloured by cohort; a curve that rises from 0% to 100% means that the selector abstains when the effect is weak and selects the true variable when it is strong. The ticks below the axis mark the RESI of each candidate batch-effect variable of the real cohorts before harmonization. (c) Outcome of the trials with a batch effect at the operating point, by effect scale and cohort: dark, true batch variable selected; grey, abstention. Selections of a wrong variable (4 trials) are too few to be visible. RESI, robust effect size index.

## Experiment 2: batch-effect removal in real cohorts

ParsiComBat chose the reconstruction kernel in NSCLC and slice thickness and manufacturer in MAMA-MIA (top three candidates in Supplementary Results). Table 3 and Figure 4 show the technical variation that each strategy left, as the mean kBET over the candidate batch-effect variables and as the number of variables that still differed between batches.

In NSCLC (Figure 4a), correcting for kernel alone was almost as effective as correcting for all 6 candidates. The mean kBET fell from 0.276 without harmonization to 0.032 with ParsiComBat and 0.021 with Greedy, and neither left any variable with a residual effect. Manual left 4 of 6 variables with a residual effect, and AutoCombat left part of the kernel effect in place.

In MAMA-MIA (Figure 4b), no strategy removed the batch effect completely. Greedy, with 7 variables, removed the most (mean kBET of 0.265; 3 of 8 variables with residual effect), followed by ParsiComBat (0.325) and Manual (0.395), which both left 6 of 8 variables; AutoCombat left the most. Only field strength kept its kBET close to its value without harmonization under every strategy. ParsiComBat thus matched Greedy in NSCLC with a single variable and removed slightly more batch effect than Manual in MAMA-MIA.

Harmonization barely changed the correlation structure of the features: the structure correlation was at least 0.979 for every strategy. Protecting sex as a covariate in NSCLC did not change the variable that ParsiComBat chose, and the other results were very similar (see Supplementary Results).

**Table 3.** Batch effects corrected and residual batch effect in Experiment 2. RESI, pseudo-F, ASW and kBET are means over the candidate batch-effect variables where lower values indicate a smaller residual batch effect. Variables with residual effect counts the candidate variables whose PERMANOVA test after harmonization had $p < 0.05$.

| Strategy | Batch variables | *n* | RESI | Pseudo-F | ASW | kBET | Structure correlation | Variables with residual effect |
|---|---|---|---|---|---|---|---|---|
| **NSCLC-Radiomics (CT)** | | | | | | | | |
| Raw | None | 0 | 0.128 | 4.17 | 0.060 | 0.276 | 1.000 | 5/6 |
| Manual | Manufacturer | 1 | 0.082 | 2.15 | 0.042 | 0.176 | 0.998 | 4/6 |
| Greedy | All feasible candidates[a] | 6 | 0.000 | 0.32 | 0.022 | 0.021 | 0.992 | 0/6 |
| ParsiComBat | Reconstruction kernel | 1 | 0.015 | 0.78 | 0.026 | 0.032 | 0.991 | 0/6 |
| AutoCombat | Clusters of all candidates[b] | 7 | 0.025 | 0.63 | 0.035 | 0.050 | 0.996 | 1/6 |
| **MAMA-MIA, ISPY2 (DCE-MRI)** | | | | | | | | |
| Raw | None | 0 | 0.197 | 21.02 | 0.104 | 0.680 | 1.000 | 8/8 |

| Strategy | Batch variables | *n* | RESI | Pseudo-F | ASW | kBET | Structure correlation | Variables with residual effect |
|---|---|---|---|---|---|---|---|---|
| Manual | Manufacturer | 1 | 0.143 | 6.99 | 0.060 | 0.395 | 0.984 | 6/8 |
| Greedy | All feasible candidates[c] | 7 | 0.046 | 2.08 | 0.042 | 0.265 | 0.979 | 3/8 |
| ParsiComBat | Slice thickness, manufacturer | 2 | 0.137 | 6.19 | 0.066 | 0.325 | 0.981 | 6/8 |
| AutoCombat | Clusters of all candidates[d] | 8 | 0.157 | 12.96 | 0.089 | 0.478 | 0.997 | 8/8 |

ASW, average silhouette width; CT, computed tomography; DCE-MRI, dynamic contrast-enhanced magnetic resonance imaging; kBET, k-nearest neighbour batch-effect test rejection rate; *n*, number of batch-effect variables used as batch variables in ComBat (for AutoCombat, the variables whose combination defined the clusters); NSCLC, non-small cell lung cancer; PERMANOVA, permutational multivariate analysis of variance; RESI, robust effect size index; Structure correlation, Spearman correlation between the feature–feature correlation matrices before and after harmonization.

[a] Greedy, NSCLC: manufacturer, scanner model, peak tube voltage, tube current, exposure, reconstruction kernel.

[b] AutoCombat, NSCLC: manufacturer, scanner model, peak tube voltage, tube current, exposure, exposure time, reconstruction kernel.

[c] Greedy, MAMA-MIA: field strength, pixel spacing, slice thickness, manufacturer, bit depth, echo time, repetition time.

[d] AutoCombat, MAMA-MIA: field strength, pixel spacing, slice thickness, manufacturer, scanner model, bit depth, echo time, repetition time.

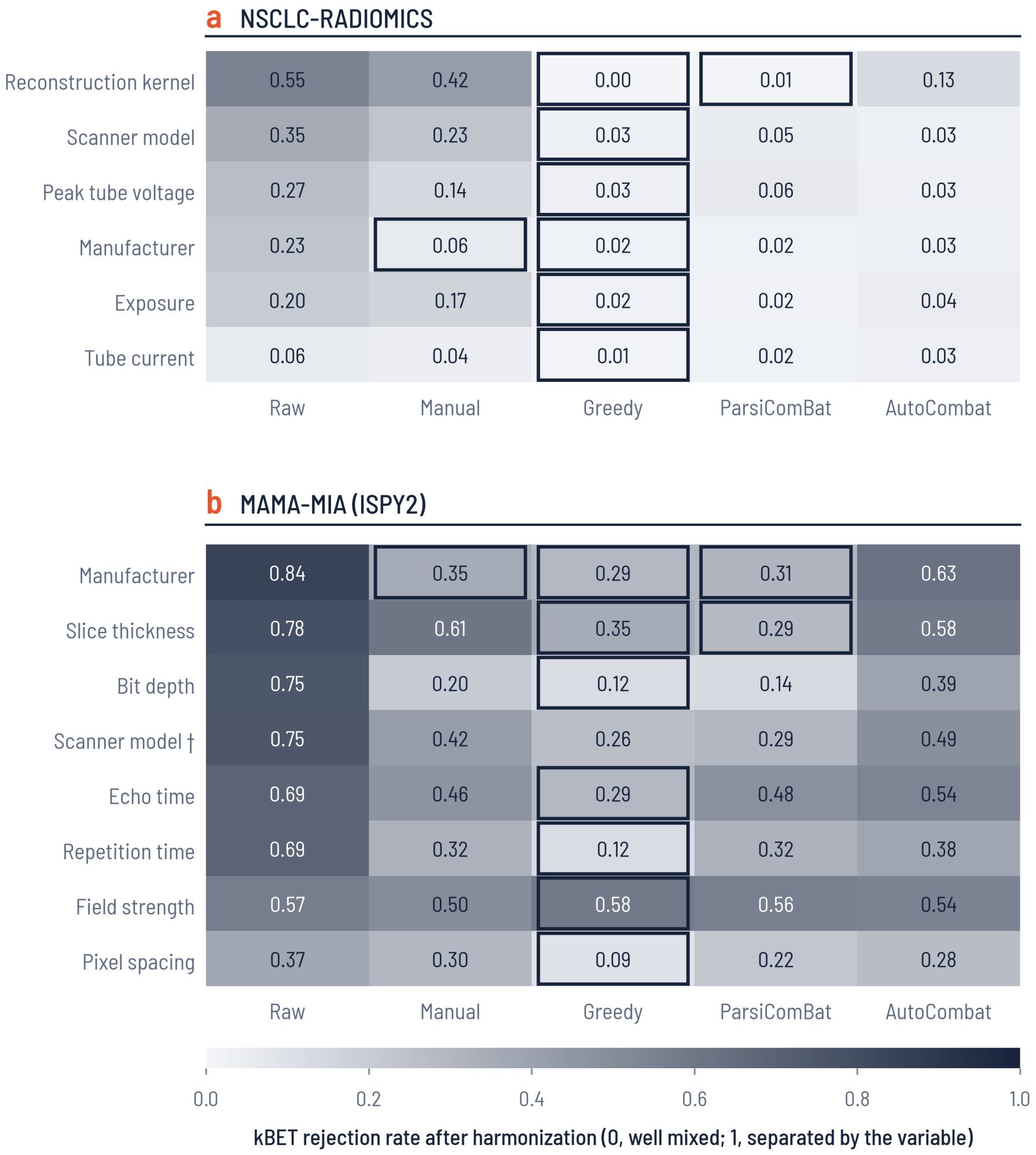

**Figure 4.** Residual batch effect of each candidate batch-effect variable after each harmonization strategy in Experiment 2. The residual batch effect was measured for all candidates, as acquisition parameters are correlated and correcting for one also reduces the effect of the others. Each cell gives the kBET rejection rate: 0 means the effect was removed and 1 that it was unchanged. Outlined cells mark the variables each strategy used as batch

variables; as AutoCombat used cluster labels, none of its cells is outlined. Comparing each column with the Raw column shows how much of the effect each strategy removed. †, scanner model could not be used by Greedy or ParsiComBat. (a) NSCLC-Radiomics. (b) MAMA-MIA, ISPY2 subset. kBET, k-nearest neighbour batch-effect test.

## Experiment 3: effect of harmonization on clinical prediction

Harmonization strategy had little effect on clinical prediction: the ROC AUC varied more between endpoints than between strategies (Table 4). In internal validation, the AUC of the five strategies was 0.71–0.75 for NSCLC stage, 0.58–0.60 for NSCLC 2-year survival and 0.67–0.70 for ISPY2 pCR. When the pCR models were applied to new cohorts, it was 0.63–0.65 in Duke, 0.66–0.70 in ISPY1 and 0.54–0.61 in NACT, where the confidence intervals were widest.

Compared with each alternative, ParsiComBat changed the AUC by −0.025 to +0.070 across the 24 comparisons (Figure 5). No confidence interval excluded zero, and no difference was significant after Holm adjustment. Balanced accuracy and F1 score led to the same conclusion, and each final internal model used 3–11 radiomic features (see Supplementary Results).

**Table 4.** Discrimination of the clinical prediction models in Experiment 3. ROC AUC of each strategy and, for each alternative strategy, the difference ParsiComBat minus that strategy on the same patients (positive differences favour ParsiComBat). Confidence intervals are paired class-stratified bootstrap intervals. Holm-adjusted $p$-values correct the paired randomization tests for the four comparisons within each endpoint or cohort.

| Endpoint | Strategy | ROC AUC (95% CI) | ΔAUC, ParsiComBat − alternative (95% CI) | Holm-adjusted *p* |
|---|---|---|---|---|
| **Internal validation (nested cross-validation)** | | | | |
| NSCLC stage | ParsiComBat | 0.721 (0.666–0.775) | — | — |
| | Raw | 0.734 (0.680–0.786) | −0.013 (−0.054 to +0.028) | 1.00 |
| | Manual | 0.715 (0.659–0.769) | +0.006 (−0.036 to +0.048) | 1.00 |
| | Greedy | 0.722 (0.668–0.775) | −0.001 (−0.042 to +0.039) | 1.00 |
| | AutoCombat | 0.746 (0.693–0.796) | −0.025 (−0.066 to +0.014) | 1.00 |
| NSCLC 2-year survival | ParsiComBat | 0.584 (0.528–0.638) | — | — |
| | Raw | 0.594 (0.537–0.649) | −0.010 (−0.064 to +0.045) | 1.00 |
| | Manual | 0.591 (0.535–0.646) | −0.008 (−0.064 to +0.049) | 1.00 |
| | Greedy | 0.576 (0.520–0.630) | +0.008 (−0.044 to +0.060) | 1.00 |
| | AutoCombat | 0.597 (0.540–0.652) | −0.014 (−0.064 to +0.039) | 1.00 |
| ISPY2 pCR | ParsiComBat | 0.680 (0.645–0.714) | — | — |
| | Raw | 0.697 (0.662–0.732) | −0.017 (−0.042 to +0.007) | 1.00 |
| | Manual | 0.672 (0.637–0.707) | +0.007 (−0.012 to +0.027) | 1.00 |
| | Greedy | 0.675 (0.640–0.711) | +0.004 (−0.016 to +0.024) | 1.00 |
| | AutoCombat | 0.689 (0.655–0.724) | −0.010 (−0.031 to +0.011) | 1.00 |
| **External validation of the ISPY2 pCR models** | | | | |
| Duke pCR | ParsiComBat | 0.652 (0.580–0.720) | — | — |
| | Raw | 0.649 (0.575–0.720) | +0.003 (−0.033 to +0.039) | 1.00 |
| | Manual | 0.651 (0.578–0.721) | +0.001 (−0.029 to +0.031) | 1.00 |
| | Greedy | 0.627 (0.553–0.700) | +0.024 (−0.004 to +0.055) | 1.00 |
| | AutoCombat | 0.643 (0.569–0.713) | +0.009 (−0.014 to +0.033) | 1.00 |
| ISPY1 pCR | ParsiComBat | 0.687 (0.601–0.768) | — | — |
| | Raw | 0.677 (0.588–0.762) | +0.010 (−0.042 to +0.061) | 1.00 |
| | Manual | 0.669 (0.579–0.755) | +0.018 (−0.016 to +0.055) | 1.00 |
| | Greedy | 0.704 (0.620–0.784) | −0.017 (−0.048 to +0.014) | 1.00 |

| Endpoint | Strategy | ROC AUC (95% CI) | ΔAUC, ParsiComBat − alternative (95% CI) | Holm-adjusted *p* |
|---|---|---|---|---|
| | AutoCombat | 0.658 (0.569–0.743) | +0.029 (−0.008 to +0.068) | 1.00 |
| NACT pCR | ParsiComBat | 0.605 (0.401–0.794) | — | — |
| | Raw | 0.585 (0.377–0.782) | +0.021 (−0.062 to +0.111) | 1.00 |
| | Manual | 0.547 (0.350–0.741) | +0.058 (−0.012 to +0.130) | 0.60 |
| | Greedy | 0.583 (0.379–0.779) | +0.022 (−0.060 to +0.096) | 1.00 |
| | AutoCombat | 0.535 (0.340–0.731) | +0.070 (−0.003 to +0.154) | 1.00 |

AUC, area under the curve; CI, confidence interval; NSCLC, non-small cell lung cancer; pCR, pathological complete response; ROC, receiver operating characteristic; —, reference strategy (no difference or *p*-value computed).

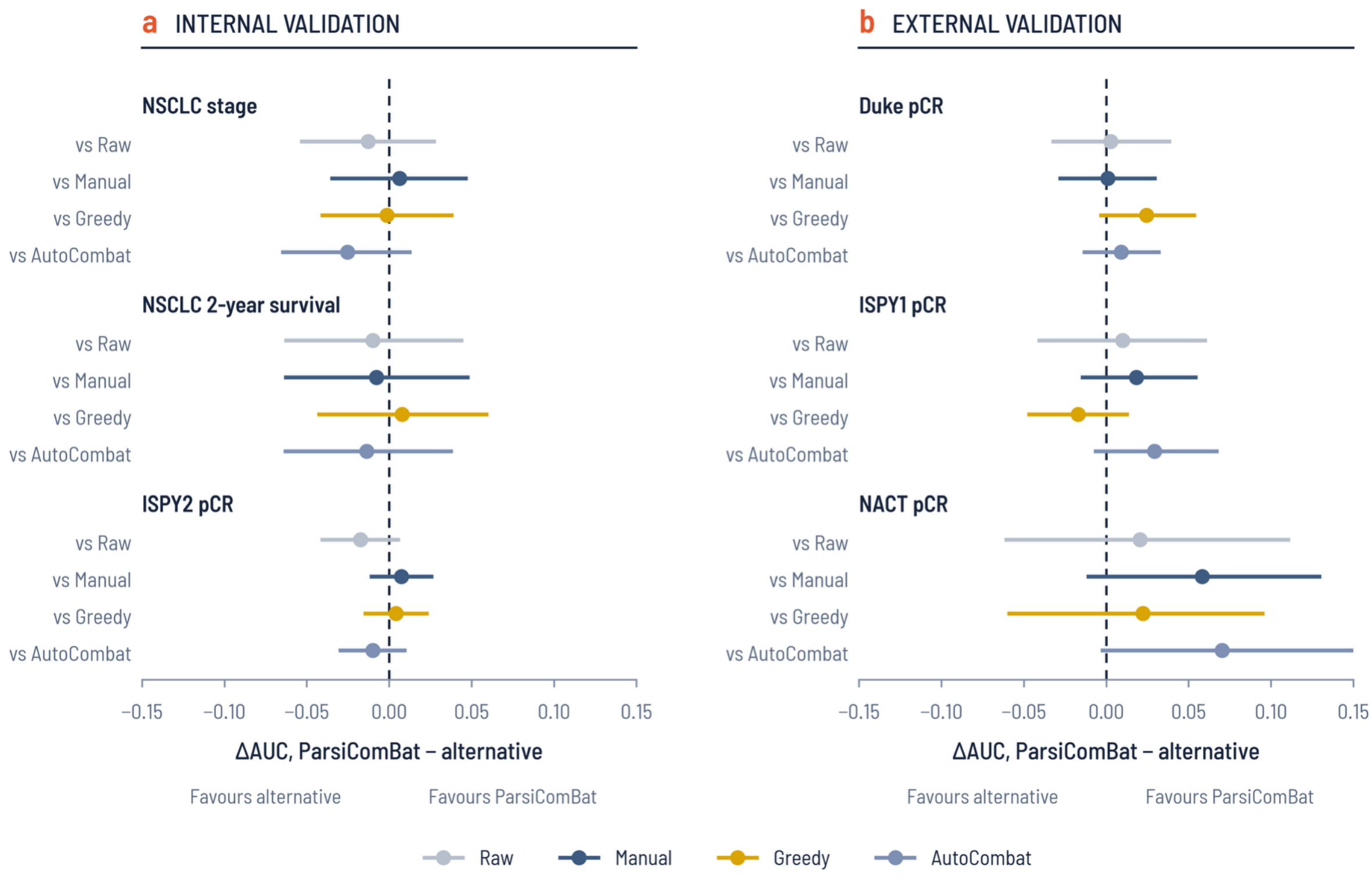

**Figure 5.** Paired differences in ROC AUC between ParsiComBat and each other strategy in Experiment 3. (a) Internal validation by nested cross-validation on the training cohort. (b) External validation of the ISPY2 pCR models, applied to the Duke, ISPY1 and NACT cohorts. Each point is the AUC of ParsiComBat minus the AUC of one alternative strategy on the same patients, coloured by strategy, and each bar its paired-bootstrap 95% confidence interval. Points right of the dashed zero-line favour ParsiComBat and a bar that crosses it is compatible with no difference. AUC, area under the curve; pCR, pathological complete response; ROC, receiver operating characteristic.

## Discussion

ComBat leaves the choice of which acquisition variable defines the batches to the user. ParsiComBat makes this choice from the data, selecting among the parameters recorded in the image headers, or none when no batch effect is large enough. In simulated data, it selected the true batch variable when the effect was at least moderate and abstained otherwise; in the real cohorts, it selected the candidates with the largest batch effect; and in clinical prediction models, these corrections did not change discrimination.

In the simulations, it identified the true batch variable in 92.9% of trials with a batch effect and abstained in 99.2% of null controls. The abstentions occurred mainly when the effect was below a RESI of 0.2, just under the conventional threshold for a medium effect (0.25) [37]. Unlike the other strategies compared, ParsiComBat can decline to harmonize. This matters because ComBat on a variable without a batch effect can introduce spurious group differences, especially when the batch is unevenly distributed across the groups of interest [17,38].

In the CT cohort, correcting for reconstruction kernel alone removed almost as much technical variation as correcting for all six feasible candidates, and more than correcting for manufacturer or using AutoCombat. Because reconstruction kernels are vendor-specific, kernel groups are nested within manufacturer, so this single correction also corrected for manufacturer. This choice agrees with phantom studies, in which the kernel affected the reproducibility of CT radiomic features more than any other acquisition parameter tested [4], although our data cannot fully separate the kernel from the scanner model. Since ParsiComBat selects specific batch variables, its choice can be checked against the acquisition protocol, which the cluster labels of AutoCombat do not allow.

In the MRI cohort, technical variation was spread over many related acquisition variables. With two corrections, ParsiComBat reduced it slightly more than correcting for manufacturer alone. Greedy removed more with seven sequential corrections and did not reduce discrimination, but this does not exclude a cost: each additional correction adds parameters estimated from the data and can introduce spurious differences [17], and the order of sequential corrections changes the result [18]. No strategy removed the effect of field strength, which suggests that field strength alters features in ways ComBat does not address [11,19].

In the clinical prediction models, ParsiComBat did not significantly change discrimination compared with no harmonization. Previous studies of ComBat in radiomic prediction models have reported gains [21,39] as well as no significant improvement, including in external validation [22,23], and in one study the variant that harmonized best predicted worst [18]. Three factors may explain this here: the final models used few radiomic features (3–11), the technical variation removed need not be associated with the endpoints, and ComBat fitted on ISPY2 cannot remove acquisition differences specific to the external centres. For ParsiComBat, this means no detectable cost, but also no evidence of a predictive gain from harmonization in these cohorts.

These results have practical implications for multicentre radiomic studies. The batch variable for ComBat should be chosen and reported on the basis of the data rather than by convention, and harmonization should not be assumed to be necessary in every cohort. Because a data-driven selector can only correct for what is recorded, acquisition parameters should be preserved in the image headers and reported.

This study has limitations. It was retrospective and used two public cohorts, one of them from a single centre, so the findings need confirmation in more centres, scanners and modalities. ParsiComBat can only select among the variables that were recorded, and the image headers were incomplete: 98 of 422 (23%) CT records had no kernel or tube voltage, and acquisition differences that were not recorded could not be corrected. Because the method selects ComBat batch variables, only ComBat-based strategies were compared; image-level and deep learning methods [8,9,40] act at a different step and were not evaluated. Finally, Experiment 3 assessed discrimination but not calibration, which ComBat can also affect [39].

## Conclusion

ParsiComBat selected one or two candidate batch-effect variables as ComBat corrections, or abstained when there was no batch effect. The choice of batch variable matters: correcting the obvious variable can leave the main batch effect in place, and correcting for every recorded variable adds corrections without a measurable benefit. By making this choice from the data, in a way that can be verified against the acquisition protocol, ParsiComBat can make harmonization in multicentre studies more reproducible and avoid unnecessary corrections. Radiomic studies should therefore report which variables they corrected for and state when they chose not to harmonize.